\pdfoutput=1

\documentclass[11pt]{article}

\usepackage{eacl2023}

\usepackage{times}
\usepackage{latexsym}

\usepackage[T1]{fontenc}

\usepackage[utf8]{inputenc}

\usepackage{microtype}

\usepackage{inconsolata}

\usepackage{amssymb}
\usepackage{amsmath}
\usepackage{url}
\usepackage{slashbox}
\usepackage{makecell}
\usepackage{subcaption}

\usepackage{textcomp}
\usepackage{graphicx}
\usepackage{tikz}
\usepackage{arydshln}
\usepackage{array} 
\usepackage{slashbox}
\usepackage{multirow}
\usepackage{float}
\usepackage{listings}
\usepackage{xcolor}
\usepackage{longtable}
\usetikzlibrary{calc}

\usepackage{tabularx}
\usepackage{booktabs}
\usepackage{colortbl}
\usepackage[normalem]{ulem}

\pgfdeclarelayer{background}

\usetikzlibrary{matrix,shapes,arrows,fit,tikzmark,calc}
\newcolumntype{.}{D{.}{.}{-1}}

\newcommand{\berttoken}[1]{\footnotesize{{\textsc #1}}}
\newcommand{\nl}[1]{\textsl{{#1}}}
\newcommand{\nlt}[1]{\textsl{{\tiny #1}}}
\newcommand{\nlutt}[1]{\textsl{{\footnotesize #1}}}
\newcommand{\sem}[1]{\sffamily\footnotesize{#1}}
\newcommand{\alab}[1]{\ttfamily\tiny{\textbf{#1}}} 
\newcommand{\semt}[1]{\sffamily\tiny{ #1}}
\newcommand{\field}[1]{\sffamily\tiny\textcolor{blue}{#1}}
\newcommand{\desc}[1]{\sffamily\tiny\textcolor{black}{#1}}

\newcolumntype{s}{>{\hsize=.001\hsize}X}
\newcolumntype{k}{>{\hsize=.98\hsize}X}

\tikzset{   
        every picture/.style={remember picture,baseline},
        every node/.style={anchor=base,align=center,outer sep=1.5pt},
        every path/.style={thick},
        }

\newcommand\marktopleft[1]{%
    \tikz[overlay,remember picture] 
        \node (marker-#1-a) at (.7em,.1em) {};%
}
\newcommand\markbottomright[1]{%
    \tikz[overlay,remember picture] 
        \node (marker-#1-b) at (2.9em,-0.9em) {};%
    \tikz[overlay,remember picture,inner sep=3pt]
        \node[draw=blue,rounded corners,fit=(marker-#1-a.north west) (marker-#1-b.south east)] {};%
}

\makeatletter
\newcommand{\thickhline}{%
    \noalign {\ifnum 0=`}\fi \hrule height 1pt
    \futurelet \reserved@a \@xhline
}
\makeatother

\title{Semantic Parsing for Conversational \\
Question Answering over Knowledge Graphs}

 \author{Laura Perez-Beltrachini$^1$, Parag Jain$^1$, Emilio Monti$^2$, Mirella Lapata$^1$ \\
 $^1$ School of Informatics, University of Edinburgh\\
 $^2$ Amazon Alexa\\
 \texttt{\{lperez,parag.jain,mlap\}@inf.ed.ac.uk,  monti@amazon.co.uk}}
 
\begin{document}
\maketitle
\begin{abstract}
  In this paper, we are interested in developing semantic parsers which
  understand natural language questions embedded in a \emph{conversation}
  with a user and \emph{ground} them to formal queries over definitions in
  a general purpose knowledge graph (KG) with very large vocabularies
   (covering thousands of concept names and relations, and millions of
  entities).  To this end, we develop a dataset where user questions are
  annotated with \textsc{Sparql} parses and system answers correspond
  to execution results thereof.  We present two different semantic
  parsing approaches and highlight the challenges of the task: dealing
  with large vocabularies, modelling conversation context, predicting
  queries with multiple entities, and generalising to new 
  questions at test time.  We hope our dataset will serve as useful
  testbed for the development of conversational semantic
  parsers.\footnote{Our dataset and models are released
  at \href{https://github.com/EdinburghNLP/SPICE}{$\mathbb{SPICE}$}.}

\end{abstract}

\section{Introduction}

Conversational information seeking is the process of acquiring
information through conversations \cite{Zamani:ea:2022}. Recent years
have seen an increasing number of applications aiming to build
conversational interfaces based on information retrieval
\cite{Radlinski:Craswell:2017} and user recommendation
\cite{Jannach:ea:2021}. The popularity of intelligent voice assistants
such as Amazon's Alexa or Apple's Siri has further stimulated research
on question answering over general purpose knowledge graphs
(e.g.,~Wikidata). Key to question answering in this context is the
ability to ground natural language onto concepts, entities, and
relations in order to produce an \emph{executable} query
(e.g.,~\textsc{Sparql}) which will retrieve an answer or
\emph{denotation} from the knowledge graph~(KG).

This grounding process, known as \emph{semantic parsing} has been
studied in the context of one or few domain-specific databases
\cite{yu-etal-2019-cosql,10.1162/tacl_a_00422,suhr-etal-2018-learning}
or without taking the conversational nature of the task into account
\cite{reddy-etal-2014-large,yih-etal-2016-value,LC-QuAD2.0,Gu2021BeyondIT}.
However, due to the complexities of the semantic parsing task, there
are no large scale datasets consisting of information seeking
conversations with executable queries against a KB. Conversational
semantic parsing over KGs requires handling very large vocabularies
covering thousands of concept names and relations, and millions of
entities rather than specialized terms consisting of hundreds of
tables and column names. Moreover, information seeking conversations
are by nature incremental involving interrelated rather than isolated
questions.

\begin{table*}[t]
\begin{minipage}{0.74\textwidth}
\renewcommand\arraystretch{1.5}
\centering
\begin{tikzpicture}[baseline=(current bounding box.center)]
\node (tbl) {
    {\footnotesize
        \begin{tabularx}{\textwidth}{@{}l@{~~}l@{~~}l@{}l@{}}
        \arrayrulecolor{gray}
        \multicolumn{2}{c}{\textbf{Utterances}} & \multicolumn{1}{c}{\textbf{        Annotations}} & \textbf{Actions and Semantic Parses} \\
        $\mathcal{T}_1$ & {\footnotesize U:} \nlutt{Which tournament did Detroit Tigers} & \field{INTENT}=\desc{Simple Question|Single Entity}\\[-1.5ex]
        & \hspace{0.39cm} \nlutt{participate in?} & \field{ENT}=[{\semt Q650855} (\nlt{Detroit Tigers})] ,\\[-1.5ex]
        && \field{REL}=[{\semt P1923} (\nlt{participating team})], \\[-1.5ex]
        & S: \nlutt{1909 World Series} & \field{TYP}=[{\semt Q500834} (\nlt{tournament})], \\[-1.5ex]
        && \field{TRIPLE}=[{\semt(Q500834,P1923,Q650855)}],  \\[-1.5ex]
        && \field{GOLD} =[{\semt Q846847} (\nlt{1909 World Series})]\\[-1ex]
        \midrule
        $\mathcal{T}_2$ & U: \nlutt{Which sports team was the champion} & \field{INTENT}=\desc{Simple Question|Single Entity|Indirect} \\[-1.5ex]
        & \hspace{0.39cm} \nlutt{of that tournament?} & \field{ENT}=[{\semt Q846847} (\nlt{1909 World Series})] , \\[-1.5ex]
        && \field{REL}=[{\semt P1346} (\nlt{winner})], \\[-1.5ex]
        & S: \nlutt{Pittsburgh Pirates} & \field{TYP}=[{\semt Q12973014} (\nlt{sports team})],  \\[-1.5ex]
        && \field{TRIPLE}=[{\semt(Q846847,P1346,Q12973014)}],  \\[-1.5ex]
        && \field{GOLD}=[{\semt Q7199360} (\nlt{Pittsburgh Pirates})]\\[-1ex]
        \midrule
        $\mathcal{T}_3$  & U: \nlutt{Does that sports team belong to} &  \field{INTENT}=\desc{Verification|2 entities, subject is indirect}\\[-1.5ex]
        & \hspace{0.39cm} \nlutt{Sacile?} & \field{ENT}=[{\semt Q653772} (\nlt{Pittsburgh Pirates}), {\semt Q53190} (\nlt{Sacile})],\\[-1.5ex] 
        && \field{REL}=[{\semt P17} (\nlt{country})], \\[-1.5ex]
        & S: \nlutt{No} & \field{TYP}=[{\semt Q15617994} (\nlt{designation admin. territorial entity})],   \\[-1.5ex]
        & & \field{TRIPLE}=[{\semt(Q653772,P17,Q53190)}], \\[-1.5ex]
        & & \field{GOLD}=[False]
        \end{tabularx}
    }
};
\end{tikzpicture} 
\end{minipage}
\hspace{-6ex}
\begin{minipage}{0.26\textwidth}
\begin{tabular}{l}
\vspace{2ex}
\hspace{-5ex}\begin{tikzpicture}[baseline=(current bounding box.east)]
\node (tbl2) {
\begin{tabular}{l}
\marktopleft{a1}
\alab{AS:}{\semt [filter\_type, find\_rev, Q650855,P1923,Q500834]} \\
\hspace{0.7ex}\alab{SP:}\hspace{-4ex}\begin{lstlisting} 
    SELECT ?x WHERE { 
        ?x wdt:P1923 wd:Q650855.
        ?x wdt:P31 wd:Q500834.}
    \end{lstlisting} \markbottomright{a1} \\
\end{tabular}};
\end{tikzpicture} \\
\vspace{2.5ex}
\hspace{-5ex}\begin{tikzpicture}[baseline=(current bounding box.east)]
\node (tbl3) {
\begin{tabular}{l}
\marktopleft{a2}
\alab{AS:}{\semt [filter\_type, find, Q846847, P1346, Q12973014]} \\
\hspace{0.7ex}\alab{SP:}\hspace{-4ex}\begin{lstlisting} 
    SELECT ?x WHERE { 
        wd:Q846847 wdt:P1346 ?x.
        ?x wdt:P31 wd:Q12973014.}
    \end{lstlisting}  \markbottomright{a2} \\
\end{tabular}};
\end{tikzpicture} \\
\vspace{-0.8ex}
\hspace{-1.7ex}\begin{tikzpicture}[baseline=(current bounding box.center)]
\node (tbl4) {
\begin{tabular}{l}
\marktopleft{a3}
\alab{AS:}{\semt [is\_in, Q53190, find, Q653772, P17]} \\
\hspace{0.7ex}\alab{SP:}\hspace{-4ex}\begin{lstlisting} 
    ASK {wd:Q653772 
         wdt:P17 wd:Q53190.}
    \end{lstlisting} \markbottomright{a3}  \\
\end{tabular}};
\end{tikzpicture} 
\end{tabular}
\end{minipage}
\vspace*{-.3cm}
\caption{Example conversations from $\mathbb{SPICE}$. The left column
  shows dialogue turns ($\mathcal{T}_{1}$--$\mathcal{T}_{3}$) with user
  (U) and system (S) utterances. The middle column shows the
  annotations provided in CSQA. Blue boxes on the right show the
  sequence of actions (AS) and corresponding \textsc{Sparql} semantic
  parses (SP).}
\label{tab:convers:conve:example}
\end{table*}

In this work, we create $\mathbb{SPICE}$, a \textbf{S}emantic
\textbf{P}ars\textbf{I}ng dataset for \textbf{C}onversational
qu\textbf{E}stion answering over Wikidata. $\mathbb{SPICE}$ consists
of user-assistant interactions where natural language questions are
paired with \textsc{Sparql} parses and answers provided by the system
correspond to \textsc{Sparql} execution results.  We derive this
dataset from CSQA \cite{csqa}, an existing benchmark originally
proposed for retrieval-based conversational question answering
\cite{irqa-survey}.  Although CSQA does not have executable queries,
it contains a large number of natural language questions and their
corresponding answers, highlighting a range of conversational
phenomena such as coreference, ellipsis, and topic change as well as
different types of questions exemplifying varying intents.

Table~\ref{tab:convers:conve:example} shows a conversation from
$\mathbb{SPICE}$ illustrating how questions (utterances on the left)
are annotated with \textsc{Sparql} queries (SP on the right blue box).
To create a large-scale dataset (197k~conversations), we develop
\textsc{Sparql} templates for different question intents; entity,
relation, and class symbols are initially under-specified and
subsequently filled automatically to generate full \textsc{Sparql}
queries.  CSQA questions have been previously associated with logical
forms generated with custom-made grammars
\cite{d2a2018,kacupaj-etal-2021-conversational,
  marion-etal-2021-structured}. As a result, semantic parsers based on
them operate with different sets of grammar rules and are not strictly
comparable, since the grammars may have different coverage and
semantics (e.g., terminal symbols may encapsulate different degrees of
execution complexity). In $\mathbb{SPICE}$, questions are represented
with \textsc{Sparql}, a standard query language for retrieving and
manipulating RDF data.\footnote{https://www.w3.org/TR/sparql11-query/}
This allows us to compare parsers developed on the dataset on an equal
footing and facilitates further extensions (e.g.,~new question
intents), without the need to redefine the grammar and its execution
engine. In an attempt to build semantic parsers which generalize to
new entities and concepts, we further create different data splits
where new  intents appear only at test time
\cite{finegan-dollak-etal-2018-improving}.



For our semantic parsing task, we establish two strong baseline models
which tackle the large vocabulary problem and the prediction of
logical forms in different ways.  The first approach
\cite{Gu2021BeyondIT} uses \emph{dynamic vocabularies} derived from KG
subgraphs for each question and a simple sequence-to-sequence
architecture to predict \emph{complete} \textsc{Sparql} queries.  The
other approach \cite{kacupaj-etal-2021-conversational} predicts
\textsc{Sparql} \emph{query templates} and then fills in entity,
relation, and type slots by means of an entity and ontology
classifier.  Our experiments reveal several shortcomings in both
approaches, such as not being able to encode large sets of KG elements
and generate the same entity several times.  Both approaches a
struggle with ellipsis, they cannot resolve coreference when the
referent appears in the conversation context beyond the previous turn,
have reduced performance on questions with multiple entities.  and
have difficulty with unseen question intents. We discuss these
challenges and outline research directions for conversational semantic
parsing.

\section{The $\mathbb{SPICE}$ Dataset}
\label{sec:dataset}

The CSQA dataset \cite{csqa} aims to facilitate the development of QA
systems that handle complex and inter-related questions over a
knowledge graph (KG).  In contrast to simple factual questions that
can be answered with a single KG triple (i.e., \{subject, relation,
object\}), complex questions require manipulating sets of triples and
reasoning over these. In Table~\ref{tab:convers:conve:example}, a
question like \nl{How many sports teams participated in that
  tournament?} requires numerical reasoning and answering the question
in turn~$\mathcal{T}_2$ relies on correctly
interpreting~$\mathcal{T}_1$.

Questions and answers in this dataset were elicited from human experts
playing user and system roles as well as from crowd-workers. In a
second stage, templates derived from the human-authored QA pairs were
used to automatically augment the dataset. Human experts also
suggested complex reasoning questions and derived templates
thereof. Conversations were built as sequences of QA pairs exploring
paths in the KG. By construction, the QA pairs in a conversation are
connected through one or several entities in the KG.  Questions fall
into two coarse categories, \textit{simple} and
\textit{reasoning}-based, and the way QA pairs are organised in a
sequence introduces various \textit{conversational phenomena} which we
summarize below.

\paragraph{Simple Questions} are factoid questions,
seeking information related to an entity (e.g., \nl{Which tournament
  did Detroit Tigers participate in?}  in
Table~\ref{tab:convers:conve:example}) or set of entities (e.g.,
\nl{What are the countries of those sports teams?}).

\paragraph{Reasoning Questions} are complex questions
which require the application of numerical and logical operators over
sets of entities.  For instance, to answer the question \nl{How many
  sports teams participated in that tournament?}  requires finding the
set of sports teams that participated in a given tournament (e.g.,
\nl{1909 World Series}) and taking its count.  Questions in this
category also involve General Entities (GE) such as \nl{tournament},
in addition to Named Entities (NE), and multiple entities (both NE and
GE) in a single question (e.g., \textsl{Which tournaments have less
  number of participating sports teams than 1909 World Series?}).
Some question types also combine multiple reasoning operators.

\paragraph{Conversations} contain sequences of
mixed-initiative interactions where the system requests clarification
on ambiguous questions.  Conversations also include discourse
phenomena such as coreference (e.g., \nl{Which sports team was the
  champion of that tournament?} in
Table~\ref{tab:convers:conve:example}) and ellipsis (e.g., \nl{And
  what about 1910 World Series?} as a follow up question for \nl{How
  many sports teams participated in that tournament?}).

There are 10 question types and 47 question sub-types.  In
Table~\ref{tab:convers-stats}, we only list question types but provide
all sub-types in Table~\ref{tab:question:tupe:subtype} in Appendix~\ref{app:dataset}.

\subsection{Question Semantics Described by Actions}

\citet{csqa} envisaged CSQA as a benchmark for retrieval-based
conversational question answering
\cite{BordesUCW15,dong-etal-2015-question,jain-2016-question,irqa-survey}.
These methods embed natural language questions and KG triples into
high dimensional spaces and rely on neural reasoning modules to match
questions to candidate answers.  Hence, questions do not have
associated logical forms, only gold answers are available.

Our success in creating semantic parse annotations is partly due to
the fact that CSQA provides useful KG information.  Each interaction
(i.e., user and system turn) comes with annotations about KG~entities,
types, and relation symbols as well as some information about the
triple patterns involved in the question (illustrated in
Table~\ref{tab:convers:conve:example} with {\small \sffamily{ENT}},
{\small \sffamily{REL}}, {\small \sffamily{TYP}}, and {\small
  \sffamily{TRIPLE}} fields).  It also provides information pertaining
to question types and sub-types (see {\small \sffamily{INTENT}} in
Table~\ref{tab:convers:conve:example}).

Taking advantage of these annotations, follow-on work \cite{d2a2018}
defined a semantic parsing task over CSQA, modeling the meaning of
questions as a sequence of actions.  The set of actions encompasses
{\sem find} (or {\sem find\_rev} when the entity is in object
position) to retrieve sets of entities in a subject (object) position,
as well as actions operating on sets of entities (e.g.,~{\sem
  filter\_type}).  For instance, the question in turn $\mathcal{T}_1$
in Table~\ref{tab:convers:conve:example} would be parsed to {\sem
  [filter\_type, find\_rev, Q650855, P1923, Q500834]}, meaning ``find
the set of entities that are in relationship \nl{participating team}
with \nl{Detroit Tigers} and then filter those that are of type
\nl{tournament}''.  A breadth-first search algorithm generates
action-grammar annotations for each question and a sequence of
grammar-actions is considered correct if upon execution it returns the
gold answer.  Subsequent work
\cite{shen-etal-2019-multi,kacupaj-etal-2021-conversational,
  marion-etal-2021-structured} expanded this action-grammar greatly
improving its coverage (i.e.,~the number of successfully annotated
questions).

\subsection{From Actions to \textsc{Sparql} Queries}

In this work, we take a step further and map CSQA natural language
questions into vanilla \textsc{Sparql} queries.
We first analysed how intent is expressed in question types and sub-types
and then manually defined \textsc{Sparql} templates for each question
sub-type. A \textsc{Sparql} template is a query with unspecified
triple patterns in the {\sem{WHERE}} clause. For instance,
the  template for the question in turn $\mathcal{T}_1$
is {\sem{\{SELECT ?x WHERE ?x RELATION ENTITY. ?x
    wdt:P31 TYPE.\}}}. We finally modified the tool provided in
\citet{kacupaj-etal-2021-conversational} to automatically instantiate
the \textsc{Sparql} templates, providing annotations for the entire dataset
(e.g., by filling missing slots  as {\sem SELECT ?x WHERE \{?x wdt:P1923 wd:Q650855. ?x wdt:P31
  wd:Q500834.\}}).

\begin{table}[t]
\centering
{\footnotesize
  \begin{tabular}{l@{\hspace{0.2cm}}c@{\hspace{6pt}}c@{}}
  \hline
  
  Nb. instances & 197K \\
  Nb. entities & 12.8M\\
  Nb. relations & 2738\\
  Nb. types & 3064\\
  Avg. turn length & 9.5\\
  Avg. entities per conversation & 7.6\\
  Avg. types per conversations & 6.5\\
  Avg. neighbourhood per turn & 181.4 triples\\
  \end{tabular}
  \begin{tabular}{l@{\hspace{0.1cm}}m{15cm}@{\hspace{1pt}}c@{}}
  \hline
Logical Reasoning, Quantitative Reasoning, Comparative\\ 
 Reasoning, Quantitative Reasoning Count, Comparative \\
  Reasoning Count, Verification, Simple Question \\
\hline 
Clarification, Coreference, Ellipsis \\
  \hline
  \end{tabular}
}
\vspace*{-.2cm}
\caption{Statistics of $\mathbb{SPICE}$ dataset (top); general
  question types (middle); linguistic phenomena (bottom).} \label{tab:convers-stats}
\end{table} 

We imported the Wikidata snapshot provided by \citet{csqa} into a KG
in a \textsc{Sparql} server (see
Appendix~\ref{app:knowledgeGraphServer} for more details) and assessed
the correctness of \textsc{Sparql} queries by executing them and
comparing results to gold answers.  For some questions the annotation
procedure did not produce a \textsc{Sparql} parse that recovered the
gold answer.  In these rare cases, we redefined the answer if it
did not affect the conversation flow or truncated the conversation up
to that point.

Table~\ref{tab:convers-stats} shows various statistics for
$\mathbb{SPICE}$ while Table~\ref{tab:comparison} compares it to
related conversational datasets such as ATIS
\cite{suhr-etal-2018-learning}, SParC \cite{yu-etal-2019-sparc}, and
CoSQL~\cite{yu-etal-2019-cosql}. As can be seen, $\mathbb{SPICE}$
contains a sizeable number of training instances, its conversations are
longer, and the semantic parsing task is real-scale.

\section{The Semantic Parsing Task}
\label{sec:semantic-parsing-task}

\begin{figure*}[t]
    \centering
    \begin{tabular}{@{}c@{\hspace*{-.55cm}}c@{}}
        \includegraphics[width=.43\linewidth,
        keepaspectratio]{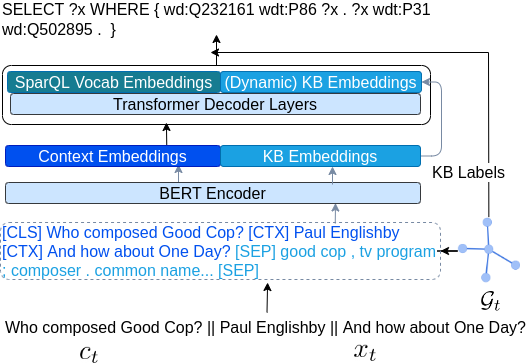} &         \includegraphics[width=.51\linewidth, keepaspectratio]{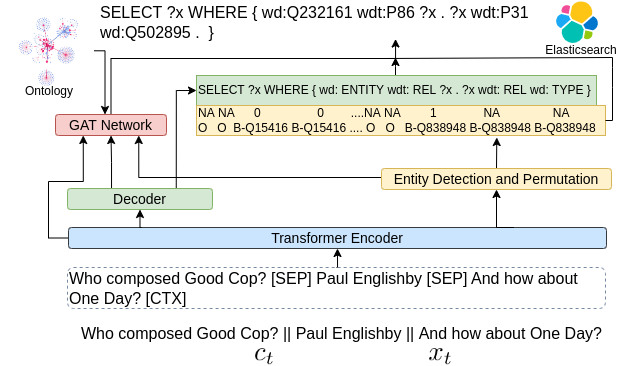} \\
(a) {\small Encoder-decoder model with dynamic vocabularies.} &
(b) {\small Semantic parser based on Lasagne multi-task model.} \\
\end{tabular}
\vspace*{-.3cm}
    \caption{Two modeling approaches to conversational semantic
      parsing.}
    \label{fig:BertSP:LasagneSP:input:format}
\end{figure*}

We consider the semantic parsing task over a sequence of dialogue
turns $d=(d_1, d_2, \cdots, d_{|d|})$, where turn~$d_t$ corresponds to
a user-system interaction with user question $x_t$ and system answer
$a_t$.  Each turn has a conversation context~$c_t$ made of
interactions $d_i$ such that~\mbox{$i<t$}.  Given interaction~$d_t$
with context~$c_t$ and user question~$x_t=(x_{t1}, x_{t2}, \cdots,
x_{t|x_t|})$, our goal is to predict a \textsc{Sparql} query
\mbox{$y_t=(y_{t1}, y_{t2}, \cdots, y_{t|y_t|})$} that represents the
intent of~$x_t$ and, upon execution over knowledge-graph~${\cal K}$,
yields denotation~$a_t$. $y_t$ is a sequence over a target vocabulary
${\cal V} = {\cal V}_f \cup {\cal V}_{{\cal K}}$ where ${\cal V}_f$ is
fixed and contains \textsc{Sparql} keywords (e.g., {\sem{SELECT}}) and
special tokens (e.g.,~beginning of sequence token, BOS), and ${\cal
  V}_{{\cal K}}$ contains all knowledge-graph symbols (e.g., entity
IDs such as {\sem{Q76}} for \nl{Barack Obama}).

We propose two approaches for this semantic parsing task which
establish strong baseline performance and highlight various
challenges. These differ in the way they handle large KG vocabularies
and how they generate logical forms.
Figure~\ref{fig:BertSP:LasagneSP:input:format} provides a sketch of the two models discussed below.

\begin{table}[t]
\centering
{\footnotesize
  \begin{tabular}{@{}l@{~~~}c@{~~}c@{~~}c@{~~}c@{}}
  \hline
     & ATIS & SParC & CoSQL & $\mathbb{SPICE}$ \\
  \hline
   Nb. Instances & 1,658 & 4,298 & 3,007 & 197K \\
   Avg. turn length & 7.0 & 3.0 & 5.2 & 9.5 \\
   Domain & Single & Multi & Multi & Wikidata \\
   Logical form & SQL & SQL & SQL & \textsc{Sparql} \\
   Database type & Rel & Rel & Rel & KG \\
  \hline
  \end{tabular}
}
\vspace*{-.2cm}
\caption{Datasets for conversational semantic parsing (Rel:~relational database; KG: knowledge graph).\label{tab:comparison}}
\end{table}

\subsection{Parsing with a Single Decoder and Knowledge Sub-graphs}
\label{sec:bertSP}

Our first model is parameterised by an encoder-decoder
Transformer neural network \cite{transformers}, and an adaptation of
the semantic parsing architecture proposed in \citet{Gu2021BeyondIT}.

\paragraph{Dynamic Vocabulary} Since the KG vocabulary~${\cal
  V}_{{\cal K}}$ can be extremely large, we parse question~$x_t$ with
a smaller vocabulary ${\cal V}_{t} \subseteq {\cal V}_{{\cal K}}$
which only contains KG symbols related to~$x_t$.  Following previous
work \cite{Gu2021BeyondIT,marion-etal-2021-structured}, we assume the
symbols related to~$x_t$ are those appearing in sub-graph ${\cal G}_t$
of knowledge-graph ${\cal K}$, ${\cal G}_t \subseteq {\cal K}$.  Given
question~$x_t$ and its context~$c_t$, we identify KG entities
\mbox{${\cal E}_t = \{e_{t1}, e_{t2}, \cdots, e_{t|{\cal E}_t|}\}$}
which correspond to mentions in~$x_t$ and~$c_t$.  We then
obtain~${\cal G}_t$ by taking the one-hop neighbourhood for each
entity~$e_{ti} \in {\cal E}_t$. In other words, we include all KG
triples ($s$, $r$, $o$) where the entity appears in subject ($s =
e_{ti}$) or object position ($o = e_{ti}$). When~$e_{ti}$ is a
subject, we include triple ($e_{ti}$, $r$, $\tau_o$) where~$\tau_o$ is
the type of entity~$o$; analogously, when~$e_{ti}$~appears in an
object position, we add ($\tau_s$, $r$, $e_{ti}$).  For entities
$e_{ti}$ we include their types $\tau_{e_{ti}}$.  When $e_{ti}$ is a
general entity (e.g., a type such as \nl{tournament}) we add 
relations from~${\cal K}$ that have instances of type~$e_{ti}$ as
their subject (object).  The final vocabulary~${\cal V}_{t}$ contains
all entities in~${\cal E}_t$, all relations~$r$ and types
($\tau_o$, $\tau_s$, and~$\tau_{e_{ti}}$) found in the set of triples
in~${\cal G}_t$.

Note that context~$c_t$ is defined as a window over the conversation
so far.  Following previous work
\cite{marion-etal-2021-structured,kacupaj-etal-2021-conversational},
we set the conversation context to the previous user-system
interaction $c_t = \{d_{t-1}\}$.

\paragraph{Encoder-Decoder Model}
Our encoder is a BERT \cite{devlin-etal-2019-bert} model fine-tuned on
our semantic parsing task. The decoder is a randomly initialised
Transformer network \cite{transformers}. 
To account for the difference 
in initialisation between the encoder and decoder networks, we follow
the training scheme proposed in \citet{liu-lapata-2019-text}.  We
provide details in Appendix~\ref{app:model}.

The input to our semantic parser is a tuple ($x_t, c_t, {\cal G}_t$)
consisting of natural language question~$x_t$, its context~$c_t$, and
subgraph ${\cal G}_t$ which we adapt to BERT's input format as follows
\cite{Gu2021BeyondIT}. We concatenate the sequence of natural language
questions and answers appearing in~$c_t$ and~$x_t$, using the special
token {\berttoken [CTX]} as a delimiter and prepend the {\berttoken
  [CLS]} token in the beginning of the sequence.  Special token
{\berttoken [SEP]} denotes the end of sequence followed by the
linearised KG sub-graph~${\cal G}_t$.  The linearisation procedure
goes over entities in~${\cal G}_t$, enumerating their types and
relations. Importantly, we denote entities by their label rather than
their KG identifiers.  The order of entities in~${\cal G}_t$ is
random.  Figure~\ref{fig:BertSP:LasagneSP:input:format}(a) shows an
example of the input to our BERT-based encoder.

More formally, the encoder takes token
sequences~$x^\prime_t~=~\text{{\berttoken
    [CLS]}}x^\prime_{t_\text{text}}\text{{\berttoken
    [SEP]}}x^\prime_{t_\text{graph}}\text{{\berttoken [SEP]}}$ as
input where $x^\prime_{t_\text{text}}$ is the natural language
subsequence 
and $x^\prime_{t_\text{graph}} = (g^t_1, \cdots, g^t_{|{\cal G}_t|})$ is
the sequence of knowledge-graph symbols from the linearised
graph~${\cal G}_t$. Note that these knowledge-graph symbols constitute
the target dynamic vocabulary ${\cal V}_t$ and $|{\cal G}_t|$
represents the number of KG symbols which is equal to the size of the
target vocabulary~$|{\cal V}_t|$.  The encoder maps input sequences
$x^\prime_t$ into sequences of continuous representations
$\mathbf{z_t} = (\mathbf{z}_{t1}, ..., \mathbf{z}_{t|x_t|})$, and the
decoder then generates the target \textsc{Sparql} parse $y_t =
(y_{t1}, ..., y_{t|y_{t}|})$ token-by-token autoregressively, hence
modelling the conditional probability: $p(y_{t1}, ...,
y_{t|y_{t}|}\,|\,x^\prime_t)$.

The linearised graph~${\cal G}_t$ can exceed BERT's maximum number of
input positions (which is 512). To avoid throwing away useful
information, we adopt a solution similar to
\citet{Gu2021BeyondIT}. For question~$x_t$ with~${\cal G}_t$
containing $k$~entities, we create $k$~input sequences $x^{\prime1}_t,
\cdots, x^{\prime k}_t$.  These $k$~sequences share the natural
language subsequence but have different KG~symbol subsequences.  Given
an input sequence $x^{\prime1}_t, \cdots, x^{\prime k}_t$, we obtain
contextualised representations as
$\mathbf{z}^1_t, \cdots, \mathbf{z}^k_t = \text{BERT}(x^{\prime1}_t, \cdots, x^{\prime k}_t)$.

The model further splits the sequence of continuous
representations~$\mathbf{z}^j_t$ into textual
representations~$\mathbf{z}^j_{t_\text{text}}$ and knowledge-graph
symbols~$\mathbf{z}^j_{t_\text{graph}}$ both of which are
contextualised.  We then average
representations~$\mathbf{z}_{t_\text{text}} =
\text{AVG}(\mathbf{z}^j_{t_\text{text}})$ and feed them as input to
the decoder (see Figure\ref{fig:BertSP:LasagneSP:input:format}(a)).
From representations~$\mathbf{z}^j_{t_\text{graph}}$, we derive the
embeddings for the elements in the target dynamic vocabulary~${\cal
  V}_t$.  Recall that the decoder parses input questions~$x_t$ using
target vocabulary ${\cal V} = {\cal V}_f \cup {\cal V}_t$ which
consists of a set of fixed (${\cal V}_f$) and dynamic (${\cal V}_t$)
target tokens.  The decoder then predicts the probability of each
\textsc{Sparql} token~$y_{ti}$ as 
$p(y_{t_i} | y_{t_{<i}}, x^{\prime1}_t, \cdots, x^{\prime k}_t ) =
\operatorname{softmax}(\mathbf{W}_o \, \mathbf{h}^L_i)$
where $\mathbf{h}^L_i$ is the decoder top layer hidden representation
at time step $i$.  $\mathbf{W}_o \in \mathbb{R}^{|{\cal V}_f \cup
  {\cal V}_t |}$ is the output embedding matrix with $\mathbf{W}_o =
[\mathbf{W}_f\,;\,\mathbf{W}_t]$, where $[;]$ denotes matrix
concatenation, and $\mathbf{W}_t$ is derived from the encoder
representations~$\mathbf{z}^j_{t_\text{graph}}$.

\subsection{Parsing with Multiple Decoders and an Ontology Classifier} 

Our second model is an adaptation of the Lasagne architecture proposed
in \citet{kacupaj-etal-2021-conversational}. 
Lasagne generates logical forms following a multi-stage approach where a backbone sketch is first predicted and then fleshed out. 
Their sketch is a sequence of actions from a custom grammar which we modify to be a sketch of \textsc{Sparql} queries. 

\paragraph{\textsc{Sparql} Template Prediction} Lasagne employs an encoder-decoder model
based on Transformers~\citep{transformers} to convert a user question
$x_t$ in a conversation into a logical form template.
The input to the encoder is the conversation context $c_t =
\{d_{t-1}\}$ and user question $x_t$. Utterances are separated via
   {\berttoken [SEP]} tokens, while the special context token
   {\berttoken [CTX]} denotes the end of sequence (see
   Figure~\ref{fig:BertSP:LasagneSP:input:format}b). The input
   sequence is encoded via multi-head attention~\cite{transformers} to
   output contextualized representations which are then fed to the
   decoder to predict a sequence of actions (without
   grounding to KG elements) token-by-token.
Instead, our decoder predicts \textsc{Sparql} queries with place-holders for KG
symbols. For instance, for the {\sem WHERE}
clause of turn $\mathcal{T}_1$ in
Table~\ref{tab:convers:conve:example}, it predicts {\sem \{ENTITY
  RELATION ?x. ?x wdt:P31 TYPE.\}} instead of {\sem \{wd:Q5582479
  wdt:P161 ?x. ?x wdt:P31 wd:Q502895.\}})

\paragraph{Entity Recognition and  Linking}
An entity recognition module detects entities in the input and links
them to the KG~\cite{shen-etal-2019-multi,
  kacupaj-etal-2021-conversational}. Initially, entity spans are
identified using an LSTM which performs BIO sequence
labelling.\footnote{BIO labels for training are obtained by preforming
  string matching between entity annotations and user utterances.}
Entity spans are subsequently linked to KG entities via an inverted
index (created using Elasticsearch\footnote{https://www.elastic.co/})
which maps entity labels to entity IDs. Once identified, the entities
are further filtered and reordered so that that they match their order
of appearance in the \textsc{Sparql} (see
Figure~\ref{fig:BertSP:LasagneSP:input:format}(b)).

\paragraph{Predicting Types and Relations} 
Finally, an ontology graph with types and
  relations appearing in $\mathbb{SPICE}$'s KG is
  constructed.\footnote{This graph would be substantially
  bigger for a semantic parsing system operating over
  the full Wikidata KG.}  The graph is encoded with a Graph Attention
Network (GAT; \citealt{DBLP:conf/iclr/VelickovicCCRLB18}) and the
prediction of type and relation fillers for the \textsc{Sparql}
template is modeled as a classification task over graph nodes, given
the conversational context and the decoder hidden state.

\paragraph{Learning} All modules outlined above are trained in a
multi-task manner, optimizing the weighted average of the following
individual losses 
    $L = \lambda_{1}L^{\rm F} + \lambda_{2}L^{\rm G} +
    \lambda_{3}L^{\rm R} + \lambda_{4}L^{\rm O}$
where $L^{\rm F}$ is the loss of the \textsc{Sparql} template decoder,
$L^{\rm G}$ is the type and relation prediction loss using the GAT
network, $L^{\rm R}$ is the entity recognition loss, and $L^{\rm O}$
the entity reordering loss (and weights~$\lambda_{1:4}$ are learned
during training).  We refer the interested reader
to~\citet{kacupaj-etal-2021-conversational} for mode details.

\begin{table*}[t]
{\scriptsize
\begin{tabular}{ll}
\begin{minipage}[t]{3.5in}

\begin{tabular}{@{}l@{~}c@{~}c@{~}c@{~}c@{~}c@{~}c@{~}c@{~}c@{}}
  \hline
&    \multicolumn{2}{c}{BertSP$_{\rm G}$} & \multicolumn{2}{c}{BertSP$_{\rm S}$} & \multicolumn{2}{c}{BertSP$_{\rm A}$} & \multicolumn{2}{c}{LasagneSP}  \\
  \hline
  Question Type & F1 & EM & F1 & EM & F1  & EM & F1  & EM \\
  \hline
  Clarification                 & 84.89 & 82.53 &  80.21 & \textbf{77.69}  &  83.91 & 76.58  &  86.29 & 73.41 \\
  Logical Reasoning (All)       & 90.61 & 82.90 &  85.55 & \textbf{66.89}  &  22.74 & 28.61  &  88.80 & 57.41 \\  
  Quantitative Reasoning (All)  & 94.42 & 88.55 &  82.95 & 66.40  &  76.20 & 59.01  &  94.90 & \textbf{91.47} \\
  Comparative Reasoning (All)   & 96.23 & 87.39 &  90.44 & 73.80  &  69.56 & 39.37  &  94.20 & \textbf{85.05} \\
  Simple Question (Coreferenced)& 88.96 & 86.53 &  83.19 &\textbf{69.87}  &  76.51 & 58.83  &  84.73 & 60.90 \\
  Simple Question (Direct)      & 91.81 & 91.59 &  87.13 & \textbf{80.69}  &  71.43 & 58.71  &  87.21 & 66.88 \\
  Simple Question (Ellipsis)    & 79.51 & 89.71 &  72.50 & \textbf{71.67}  &  58.14 & 50.90  &  74.35 & 61.53 \\
  \hline
   & AC & EM & AC & EM & AC & EM & AC  & EM \\
  \hline
  Verification (Boolean)        & 90.10 & 77.24 &  79.72 & \textbf{62.62}  &  37.16 & 24.90  &  34.89 & 26.72 \\
  Quantitative Reasoning (Count)& 87.91 & 84.97 &  76.88 & \textbf{73.20}  &  50.86 & 48.44  &  60.51 & 56.15 \\
  Comparative Reasoning (Count) & 90.05 & 85.99 &  73.18 & 66.79  &  43.48 & 40.67  &  89.09 & \textbf{83.69} \\
  \hline
  \hline
  Overall & 81.50 & 85.74 & 81.18 & \textbf{70.96} & 59.00 & 48.60 & 79.50 & 66.32 \\
  \hline
  \end{tabular}

\vspace*{-.2cm}
\caption{Accuracy (AC), and exact match (EM) on $\mathbb{SPICE}$ i.i.d test split. BertSP$_{\rm G}$ has access to
  oracle entities, types, and coreference annotations. Best EM
  predictions are shown in bold.}
\label{tab:convers:iidtest:models}
\end{minipage}
&
\raisebox{1.3cm}[0pt]{\begin{minipage}[t]{2.5in}
  \begin{tabular}{l@{~}|c@{~}c@{~}c@{~}c@{}}
  \hline
   Phenomena & BertSP$_{\rm G}$ & BertSP$_{\rm S}$ & BertSP$_{\rm A}$ & LasagneSP  \\
  \hline
  Coreference$_{=-1}$    & 81.40 & 70.65 & 49.39 & 43.65 \\
  Coreference$_{<-1}$    & 67.82 & 0 & 0 & 0 \\
  Ellipsis               & 75.93 & 54.33 & 26.39 & 46.54 \\
  Multiple Entities      & 83.37 & 65.40 & 41.64 & 66.52 \\
  \hline
  \end{tabular}
\vspace*{-.2cm}
\caption{Exact match on $\mathbb{SPICE}$ i.i.d test;
  questions grouped into linguistic phenomena. 
  \label{tab:convers:iidtest:cases}}
\end{minipage}} \\
&
\raisebox{2cm}[0pt]
{\begin{minipage}[t]{2.5in}
  \begin{tabular}{@{}l@{~~}c@{}c@{~}c@{}}
  \hline
Unseen   & Instances & BertSP$_{\rm S}$ & LasagneSP \\
   Combinations  & Train/valid/test & EM & EM \\
  \hline
   \textsc{CountLogic}       & 153,562/14,262/29,177 & ~~0.94 & 0 \\ 
   \textsc{UnionMulti}       &  157,331/14,426/25,244 & 19.74  &  16.89 \\ 
   \textsc{Verify3}          & 154,027/13,869/29,105  & 0  &  0 \\ 
  \hline
  \end{tabular}
\vspace*{-.2cm}
\caption{Exact match for BertSP$_{\rm S}$ and
  LasagneSP on $\mathbb{SPICE}$ non-i.i.d
  splits. \label{tab:csqaTSplitsResults}}
\end{minipage}}
\end{tabular}
}
\end{table*} 

\section{Results}
\label{sec:results}

We examine how the two models just described fare on different
question types and subtypes. We report results on $\mathbb{SPICE}$
i.i.d train/valid/test splits (containing 152,391/16,813/27,797
conversations, respectively) but also create new splits that assess
out-of-distribution generalization.  In all cases, following previous
work \cite{csqa,kacupaj-etal-2021-conversational}, we use
execution-based automatic metrics. Micro \emph{F1-score} evaluates question
parses that return a set of entities, while \textit{Accuracy} is used
for question parses that evaluate to True/False or return a numerical
value.
In addition, we report \textit{Exact Match} (EM) against
the gold \textsc{Sparql} parse.

\subsection{Performance per Question Type}
\label{sec:results:qtypes}

Table~\ref{tab:convers:iidtest:models} shows our results on the
$\mathbb{SPICE}$ i.i.d test~split.
BertSP$_{\rm G}$ has access to oracle entities, types, and coreference
annotations which allows us to disassociate the complexity of the
\textsc{Sparql} generation task from the problem of grounding and
disambiguating entities to KG symbols.  Variants BertSP$_{\rm
  S}$ and BertSP$_{\rm A}$ do not have access to oracle
annotations.  BertSP$_{\rm S}$ grounds mentions to KG
entities with a simple algorithm based on string matching
\cite{marion-etal-2021-structured}; while BertSP$_{\rm A}$ relies on
 AllenNLP's Named Entity Recognizer (NER) 
 and the Elasticsearch inverted index for Named Entity Linking (NEL).  
Both have to identify coreferring entities using the conversation context~$c_t$.
Both BertSP$_{\rm S}$ and BertSP$_{\rm A}$ use string matching for
type linking (i.e.,~grounding general entities to KG symbols).

Note that it is not straightforward to perform oracle analysis for
LasagneSP without compromising the model structure which predicts
entities, their types, and relations in multiple stages.

\paragraph{Exact Match Performance} We observe that execution based metrics (\mbox{F1-score} 
and Accuracy) are generally higher than~EM. This is because in some
cases the \textsc{Sparql} parse may be incorrect and still yield some
results. For instance, a parse requiring the {\small \textsc{UNION}}
of two graph patterns may yield a partially correct answer by only
including one graph pattern; similarly, a parse can evaluate to False
and agree with the gold answer just because it included a wrong
relation symbol.

\paragraph{The Importance of Entity Grounding} 
Not surprisingly, the model with access to oracle information (variant
BertSP$_{\rm G}$) obtains the best performance. Results improve not
only for questions with entities referring to previous context but
also indirectly for other types of questions. Since entities are
correctly grounded in previous conversation turns~$c_t$, the model
operates with more accurate~graphs ${\cal G}_t$ and richer dynamic
vocabularies~${\cal V}_t$.

Both BertSP$_{\rm S}$ and BertSP$_{\rm A}$ perform coreference
resolution using limited conversation context and thus performance
decreases.  These models also have to ground named (\nl{Detroit
  Tigers}) and general (\nl{tournament}) entity mentions to
KB symbols. BertSP$_{\rm S}$ which relies on string matching performs
overall better than  BertSP$_{\rm A}$ which struggles with  compound named entities such as
\nl{President of the Czech Republic}) and disambiguation during NEL
(e.g., \nl{Saint Barbara} the painting versus the Saint).

\paragraph{Model Comparison} 
BertSP$_{\rm S}$ and LasagneSP are similar in they way they handle
NER/NEL with a task-specific approach, but differ in their
conceptualization of the semantic parsing task (encoder-decoder
vs. multi-tasking).  LasagneSP outperforms BertSP$_{\rm S}$ in
Comparative, Quantitative, and Comparative-Count questions.  These
encompass many question sub-types with general entities which are very
common in both training and testing. LasagneSP has access to
\emph{all} types and relations encoded with the graph network.  In
contrast, BertSP$_{\rm S}$ relies on types which in the first place
need to be present in the entity neighbourhood sub-graph and then be
preserved after truncating the input to fit the model's maximum
sequence length. An advantage of BertSP$_{\rm S}$ over LasagneSP, is
that it allows for easier adaptation to new types and relations by
relying on dynamic vocabularies, while  LasagneSP would need to be retrained
to accommodate them.

  In Simple questions, where each question involves fewer but more
  diverse types, BertSP$_{\rm S}$ predicts more accurate types (thanks
  to the input text and KG symbol contextualisation) and thus performs
  better.  LasagneSP does poorly on Verification, Logical, and
  Quantitative-Count (which includes logical operators).  This can be
  explained by a modelling limitation, i.e., it is not able to point
  to the same input entity more than once.

\paragraph{Errors in Predicted \textsc{Sparql}s}
Manual inspection of \textsc{Sparql} predictions revealed several
common system errors including: prediction of erroneous entities and
relations, failure to enumerate all required entities (for questions
with multiple entities), and mistakes in argument order (i.e.,
entities and variables are correct but placed in incorrect
subject/object positions). To a lesser extent, we also observed
\textsc{Sparql} queries with incorrect intent predictions and
ill-formed syntax.

\subsection{Linguistic Analysis}

Table~\ref{tab:convers:iidtest:cases} shows model performance
across-different question sub-types aggregated for specific phenomena.
These include coreference, ellipsis, and multiple entities. We
distinguish between cases where coreference can be resolved in the
previous turn ($d_{t-1}$) and further back in the conversation history
($d_{t-i}$, where $i>1$ ).  In addition, some question sub-types
contain plural mentions, i.e., they are linked to multiple entities
which the semantic parser must enumerate in order to build the correct
parse.  Ellipsis can be often resolved within the previous interaction
($d_{t-1}$).  Questions with multiple entities bring further
disambiguation challenges. In Appendix~\ref{app:dataset}, we provide
the list of question sub-types for each phenomenon in
Table~\ref{tab:convers:iidtest:cases}.

As can be seen, the oracle BertSP$_{\rm G}$ model which has access to
gold annotations is superior to variants which rely on automatic
entity and type linking. BertSP$_{\rm S}$ is better than LasagneSP at
handling coreference within immediate context (i.e.,~$c_t =
\{d_{t-1}\}$).  Due to the fact that LasagneSP predicts entity
positions in \textsc{Sparql}, it is particularly bad at resolving
mentions to multiple entities in the previous context or even multiple
mentions of the same entity in the output parse (as is the case with
Verification questions).  Perhaps unsurprisingly, neither BertSP$_{\rm
  S}$ nor LasagneSP can resolve mentions to non-immediately preceding
utterances.  BertSP$_{\rm S}$ performs better than LasagneSP in
questions with ellipsis; we conjecture that the input context and
contextualisation of KG symbols help in grounding elided relation
mentions.  Ellipsis and multiple entities improve by a large margin
with access to gold annotations (see BertSP$_{\rm G}$ in
Table~\ref{tab:convers:iidtest:cases}).

\subsection{Generalisation}

We further evaluate the models' ability to generalize by creating
``query-based'' splits \cite{finegan-dollak-etal-2018-improving},
i.e., splits with query patterns seen only at test time.  Our
splits, shown in Table~\ref{tab:csqaTSplitsResults},
include: (a) question sub-types that involve a count operation
over a union operator (\textsc{CountLogic}; individual operators are
seen at training time but not the combination thereof); (b)~question
sub-types that involve a union operator over two graph patterns with
different relations (\textsc{UnionMulti}; the union of two graph
patterns with the \emph{same} relation is seen during training); and
(c)~verification questions with three entities (\textsc{Verify3};
questions with 2 entities only are seen during training).

As shown in Table~\ref{tab:csqaTSplitsResults}, both BertSP$_{\rm S}$
and LasagneSP perform poorly across different splits.
While in some cases the models grasp the overall \textsc{Sparql}
structure for unseen questions (e.g.,~\nl{Which watercourses are
  located in the neighbourhood of Bremen or are the tributaries of Ob?
  in \textsc{UnionMulti}}), they ignore specific query details and
simply default to familiar patterns seen in the training (e.g.,
\nl{Which people are the creators of The Theory of Everything or Ten
  Minutes to Live?}). In the \textsc{UnionMulti} split, the models
produce an appropriate \textsc{Sparql} template but systematically
copy the \emph{same} relation in \emph{both} graph patterns.
BertSP$_{\rm S}$ performs slightly better than LasagneSP; we
hypothesize that  contextualised KG embeddings
occasionally help the model select different relations. We observe a
similar trend for \textsc{CountLogic} and \textsc{Verify3}.  In
Appendix~\ref{app:example:output} shows examples of unseen
questions, their \textsc{Sparql}s, and common errors.

\section{Related Work}
Much previous work on semantic parsing focuses on mapping stand-alone
utterances to logical forms. Relatively few datasets have been
constructed for \emph{conversational}
semantic-parsing~\citep{suhr-etal-2018-learning,dahl-etal-1994-expanding,yu-etal-2019-sparc,yu-etal-2019-cosql}
partly due to the difficulty of eliciting annotations in an
interactive context. As a result, existing benchmarks are either
single-domain or small-scale (see the comparison in
Table~\ref{tab:comparison}). For instance, although
ATIS~\citep{suhr-etal-2018-learning} exemplifies several challenging
long-range discourse phenomena~\citep{10.1162/tacl_a_00422}, it is
restricted to a single domain with a simple database
schema. SParC~\citep{yu-etal-2019-sparc} and
CoSQL~\citep{yu-etal-2019-cosql} present cross-domain challenges in
mapping natural language queries onto SQL, but the conversation length
is fairly short and the databases relatively small-scale.

Large KGs, like Wikidata~\citep{10.1145/2187980.2188242}, are becoming
an increasing valuable source of information. Various
question-answering datasets have been recently released
(\citealt{LC-QuAD2.0,talmor-berant-2018-web}, \emph{inter alia}), but
do not cover conversational queries. The CSQA dataset introduced in
\citet{csqa} is conversational and covers a wide range of linguistic
phenomena (e.g., ellipsis, coreference) but frames the QA task as an
information retrieval problem. Follow-on
work~\citep{marion-etal-2021-structured,
  kacupaj-etal-2021-conversational,shen-etal-2019-multi} has used  hand-crafted grammars to
automatically obtain semantic annotations which are 
 are not 
executable with a real KG engine (e.g., Blazegraph), 
and cumbersome to extend to new question intents. 

\section{Conclusion}
In this work we introduce $\mathbb{SPICE}$, a conversational semantic
parsing dataset over knowledge graphs. Our dataset contains
\textsc{Sparql} annotations which are executable on a real KG engine
and requires handling complex questions, type, relation, and entity
linking on a large scale. Moreover, it showcases multiple linguistic
phenomena such as coreference and ellipsis. We establish two strong
baselines for the semantic parsing task and present detailed analysis
stratifying performance by question type and linguistic phenomena. We
also study generalization to unseen  intents and create multiple 
dataset splits with different query patterns. We hope our dataset will 
serve as a useful testbed for the development of conversational semantic
parsers.

\section{Limitations}
Both models discussed in this work make simplifying
assumptions. BertSP variants need to truncate the linearised graphs
for computational cost reasons.
LasagneSP works with a smaller
graph ontology which can easily fit in memory. However, this restrics
the model to predicting seen types or relations which is
unrealistic. A real-world semantic parser should ideally have access
to the full Wikidata . Our results show that both models do not
generalize well to unseen question intents, which is a known limitation
of current neural sequence-to-sequence
architectures~\cite{49343,finegan-dollak-etal-2018-improving,keysers2020measuring,kim-linzen-2020-cogs,li-etal-2021-compositional}. Finally, our results also suggest that there is scope for improvement in
handling previous context (including questions and answers).

\bibliography{seqspar2,anthology}
\bibliographystyle{acl_natbib}

\appendix

\section{The $\mathbb{SPICE}$ Dataset: Question Types and Sub-Types}
\label{app:dataset}

Table~\ref{tab:question:tupe:subtype} provides the list of question
types and sub-types in $\mathbb{SPICE}$. For each question sub-type
we provide an example user question. For cases involving ellipsis
and coreference, we include the conversation context (in grey colour).

Table~\ref{tab:subTypes:phenomena} provides the list of question
sub-types grouped per linguistic phenomena. Coreference ($_{=-1}$ and $_{<-1}$), Ellipsis, and Multiple Entities.

\begin{table}[h]
\centering
{\footnotesize
 \begin{tabular}{m{7cm}}
    \hline
   \multicolumn{1}{c}{Coreference} \\[1.1ex]
   More/Less | Mult. entity type (Coreference) \# 
   More/Less | Single entity type (Coreference) \# 
   Single Entity (Coreference) \#
   2 entities, one direct and one indirect, object is indirect \#
   2 entities, one direct and one indirect, subject is indirect \#
   3 entities, 2 direct, 2(direct) are query entities, subject is corefered \# 
   one entity, multiple entities (as object) corefered \#
   Count | Logical operators (Coreference) \# 
   Count | Single entity type (Coreference) \# 
   Count over More/Less | Mult. entity type (Coreference) \#
   Count over More/Less | Single entity type (Coreference) \\
   \hline
   \multicolumn{1}{c}{Ellipsis} \\[1.1ex]
   Difference | Single Relation (Ellipsis) \#
   Intersection | Single Relation (Ellipsis) \#
   Union | Single Relation (Ellipsis) \#
   More/Less | Mult. entity type (Ellipsis) \#
   More/Less | Single entity type (Ellipsis) \#
   object parent is changed, subject and predicate remain same \#
   Incomplete count-based ques \#
   Count over More/Less | Mult. entity type (Ellipsis) \#
   Count over More/Less | Single entity type (Ellipsis) \\
   \hline
   \multicolumn{1}{c}{Multiple Entities} \\[1.1ex]
   Difference | Multiple Relation \#
   Intersection | Multiple Relation \#
   Union | Multiple Relation \#
   Atleast/ Atmost/ Approx. the same/Equal | Mult. entity type \#
   Min/Max | Mult. entity type \#
   More/Less | Mult. entity type \#
   More/Less | Mult. entity type (Ellipsis) \#
   More/Less | Mult. entity type (Coreference) \#
   Mult. Entity (Simple Question Direct and Coreference) \#
   one entity, multiple entities (as object) coreferred \#
   Count over Atleast/ Atmost/ Approx. the same / Equal | Mult. entity type \#
   Count | Mult. entity type \#
   Count over More/Less | Mult. entity type \#
   Count over More/Less | Mult. entity type (Ellipsis) \#
   Count over More/Less | Mult. entity type (Coreference) \\
  \hline
  \end{tabular}
}
\caption{Question sub-types per linguistic phenomena.}
\label{tab:subTypes:phenomena}
\end{table} 

Table~\ref{tab:subTypes:noniid:splits} provides the list of
unseen question sub-types for each of the non-i.i.d splits.

\begin{table}[h]
\centering
{\footnotesize
 \begin{tabular}{m{7cm}}
    \hline
   \multicolumn{1}{c}{\textsc{CountLogic}} \\[1.1ex]
   Count | Logical operators \# Count | Logical operators (Coreference) \\
   \hline
   \multicolumn{1}{c}{\textsc{UnionMulti}} \\[1.1ex]
   Union | Multiple Relation \\
   \hline
   \multicolumn{1}{c}{\textsc{Verify3}} \\[1.1ex]
   3 entities, 2 direct, 2(direct) are query entities, subject is indirect \# 3 entities, all direct, 2 are query entities \\
   \hline
  \end{tabular}
}
\caption{Unseen question sub-types in $\mathbb{SPICE}$ non-i.i.d splits.}
\label{tab:subTypes:noniid:splits}
\end{table} 

\section{Creating a Knowledge Graph from the CSQA Data}
\label{app:knowledgeGraphServer}

Deploying a full copy of Wikidata locally as a standalone service
requires huge resources in addition to cluster dependent tweaking to
obtain fast query processing and
high-performance.\footnote{\url{https://www.mediawiki.org/wiki/Wikidata\_Query\_Service/User\_Manual\#Standalone\_service}}
To enable easier deployment and fast access for research purposes we
created a smaller graph from the CSQA data files. We mapped the
contents of these files\footnote{Described at
  \url{https://amritasaha1812.github.io/CSQA/download\_CQA/}} onto
triples which we subsequently converted to \texttt{ttl}
format\footnote{http://www.w3.org/TR/turtle/} with full URI to allow
loading them to the KG query engine. We also filled  missing
information where possible, for example, missing relations such as
``instance of'' was filled with relation~\textsc{P31} and added data type
information when this was omitted from the original files.

We used Blazegraph\footnote{https://blazegraph.com/} to deploy the
local server, which uses only CPU-based resources and has access to
150G of RAM. Further details along with the script to host the server
will be released upon acceptance.

\section{BertSP Model Configuration}
\label{app:model}

Our model is implemented using
pytorch~\citep{NEURIPS2019_bdbca288}. For all experiments, we used the
ADAM optimizer~\citep{kingma2019method} with 20,000 BERT warmup steps
and 10,000 steps for decoder warm up. We use separate optimizers for
the BERT encoder and decoder. BERT was fine-tuned during training with
the initial learning rate set to~0.00002. A learning rate of~0.001 was
set for the rest of model parameters. Our model was trained for
100,000 steps; we used 4 GPUs with 12GB of memory.  
We performed model selection on the validation set. We
report results with the best performing model which had 6 decoder
layers.

\section{Examples on Generalisation Splits}
\label{app:example:output}

Table~\ref{tab:generalisation:errors} shows examples from the generalisation
splits: similar question sub-types see during training, unseen question sub-type, and error on prediction. The most common error across different splits
is that models use similar \textsc{Sparql}s seen during training but fail to
adapt them to the details (entities, types, relations, argument positions) in the unseen question sub-type. Other errors encompass using the incorrect
\textsc{Sparql} query (incorrect question intent) and incorrect entities and
types.

\onecolumn

    {\scriptsize
\begin{center}
\begin{longtable}{lm{8.1cm}}
  \caption{Full list of question sub-types (intents) in
    $\mathbb{SPICE}$. For each subt-type we show an example user
    question, whenever the question sub-type involves a conversational
    phenomenon (coreference or ellipsis); previous
    conversation interactions necessary for the interpretation of the user question
    are provided in grey.}\label{tab:question:tupe:subtype}\\
\hline
Question Sub-Type & Example User Question \\ 
\hline
\endfirsthead
\multicolumn{2}{c}%
{\tablename\ \thetable\ -- \textit{Continued from previous page}} \\
\hline
Question Type/Sub-Type & Example User Question \\ 
\hline
\endhead
\hline \multicolumn{2}{r}{\textit{Continued on next page}} \\
\endfoot
\hline
\endlastfoot
\textbf{Clarification}     \\
\multirow{3}{*}{Simple Question | Single Entity (Coreference)} & {\tiny U:} \nlt{Which political territory is that sporting event located in?} \\
& {\tiny S:} \nlt{Did you mean Speed skating at the 2010 Winter Olympics – Men's 500 metres?} \\
& {\tiny U:} \nlt{Yes}\\[2ex]
\textbf{Logical Reasoning (All)}             \\
Difference | Multiple Relation & {\tiny U:} \nlt{Which people were awarded with Order of Merit for Arts and Science and are not working as singer?} \\
\arrayrulecolor{gray!20}\hline
Difference | Single Relation & {\tiny U:} \nlt{Which international organizations had Poland but not Bulgaria as their member?} \\
\arrayrulecolor{gray!20}\hline
\multirow{5}{*}{Difference | Single Relation (Ellipsis)} &  \textcolor{darkgray!70}{{\tiny U:} \nlt{Which city was Pierre Laffont born in?}} \\
        & \textcolor{darkgray!70}{{\tiny S:} \nlt{Marseille}} \\
        & \textcolor{darkgray!70}{{\tiny U:} \nlt{Which administrative territories are the sister cities of that city?}} \\
        & \textcolor{darkgray!70}{{\tiny S:} \nlt{Shanghai, Odessa, Naples}} \\
        & {\tiny U:} \nlt{But not Bologna} \\
\arrayrulecolor{gray!20}\hline
Intersection | Multiple Relation & {\tiny U:} \nlt{Which human settlements are situated close to Trave and have an adjacent border with Herzogtum Lauenburg?} \\
\arrayrulecolor{gray!20}\hline
Intersection | Single Relation & {\tiny U:} \nlt{Which works of art were filmed at Edinburgh and Berlin?} \\
\arrayrulecolor{gray!20}\hline
\multirow{3}{*}{Intersection | Single Relation (Ellipsis)} & \textcolor{darkgray!70}{{\tiny U:} \nlt{Which language does José María Lassalle speak in?}} \\
& \textcolor{darkgray!70}{{\tiny S:} \nlt{Spanish}} \\
& {\tiny U:} \nlt{And also Sergio Gil} \\
\arrayrulecolor{gray!20}\hline
Union | Multiple Relation & {\tiny U:} \nlt{Which watercourses are located in the neighbourhood of Bremen or are the tributaries of Ob?} \\
\arrayrulecolor{gray!20}\hline
Union | Single Relation & {\tiny U:} \nlt{Which people are the creators of The Theory of Everything or Ten Minutes to Live?} \\
\arrayrulecolor{gray!20}\hline
Union | Single Relation (Ellipsis) & \textcolor{darkgray!70}{{\tiny U:} \nlt{What is the profession of Mai Yamada?}}\\
     & \textcolor{darkgray!70}{{\tiny S:} \nlt{announcer}} \\
     & {\tiny U:} \nlt{Or Kazimierz Rogoyski?} \\[2ex]
\textbf{Quantitative Reasoning (All)}\\
Min/Max | Single entity type & {\tiny U:} \nlt{Which musical instruments are played by min number of people?} \\
\arrayrulecolor{gray!20}\hline
Min/Max | Mult. entity type & {\tiny U:} \nlt{Which organizations are the main building contractors of max number of architectural structures and buildings?}\\
\arrayrulecolor{gray!20}\hline
Atleast/ Atmost/ Approx. the same/Equal | Single entity type & {\tiny U:} \nlt{Which musical instruments are played by exactly 5 people?}\\
\arrayrulecolor{gray!20}\hline
Atleast/ Atmost/ Approx. the same/Equal | Mult. entity type & {\tiny U:} \nlt{Which events are demonstrated in atleast 3 prints and genres of sculpture?} \\[2ex]
\textbf{Comparative Reasoning (All)}   	\\
More/Less | Mult. entity type & {\tiny U:} \nlt{Which landforms are known for containing lesser number of chemical compounds or minerals naturally than Stetind pegmatite?} \\
\arrayrulecolor{gray!20}\hline
More/Less | Mult. entity type (Ellipsis) & \textcolor{darkgray!70}{{\tiny U:} \nlt{Which landforms are known for containing lesser number of chemical compounds or minerals naturally than Stetind pegmatite?}} \\
& \textcolor{darkgray!70}{{\tiny S:} \nlt{Euboea, Izalco, Mount Nyiragongo}} \\
& {\tiny U:} \nlt{And also tell me about Tuften quarry?} \\
\arrayrulecolor{gray!20}\hline
More/Less | Mult. entity type (Coreference) & \textcolor{darkgray!70}{{\tiny U:} \nlt{Which administrative territory is that person a civilian of?}} \\
& \textcolor{darkgray!70}{{\tiny S:} \nlt{Spain}} \\
& {\tiny U:} \nlt{Which administrative territories are the countries of origin of lesser number of television programs or works of art than that administrative territory?} \\
\arrayrulecolor{gray!20}\hline
More/Less | Single entity type & {\tiny U:} \nlt{Which television programs have been dubbed by more number of people than Puss in Boots: The Three Diablos?}\\
\arrayrulecolor{gray!20}\hline
More/Less | Single entity type (Ellipsis) & \textcolor{darkgray!70}{{\tiny U:} \nlt{Which television programs have been dubbed by more number of people than Puss in Boots: The Three Diablos?}} \\
& {\tiny S:} \nlt{House, K-On!, K-On!!} \\
& {\tiny U:} \nlt{And also tell me about Chip 'n Dale Rescue Rangers?} \\
\arrayrulecolor{gray!20}\hline
More/Less | Single entity type (Coreference) & \textcolor{darkgray!70}{{\tiny U:} \nlt{Which languages are max number of literary works composed in?}} \\
& \textcolor{darkgray!70}{{\tiny S:} \nlt{English}} \\
& {\tiny U:} \nlt{Which languages are the mother tongues of less number of people than that language?} \\[2ex]
\textbf{Simple Question (Direct) }   \\
Simple Question & {\tiny U:} \nlt{Which type of sport did Amel Tuka participate in?} \\
\arrayrulecolor{gray!20}\hline
Single Entity & {\tiny U:} \nlt{What is the capital of Sweden?} \\
\arrayrulecolor{gray!20}\hline
Mult. Entity & {\tiny U:} \nlt{Who were the writers of On being and essence, De vegetabilis et plantis libri septem and Historia de regibus Gothorum, Vandalorum et Suevorum?} \\[2ex]
\textbf{Simple Question (Ellipsis)}   \\
only subject is changed, parent and relation remains same & \textcolor{darkgray!70}{{\tiny U:} \nlt{Which organizations are the sponsors of Janice Anderson?}}\\
& \textcolor{darkgray!70}{{\tiny S:} \nlt{Montrail, Patagonia, Inc.}}\\
& {\tiny U:} \nlt{And also tell me about Manikala Rai?} \\
\arrayrulecolor{gray!20}\hline
object parent is changed, subject and relation remain same & \textcolor{darkgray!70}{{\tiny U:} \nlt{Which watercourses are situated nearby Munich?}} \\
& \textcolor{darkgray!70}{{\tiny S:} \nlt{Eisbach, Würm, Isar}} \\
& {\tiny U:} \nlt{And which river?} \\[2ex]
\textbf{Simple Question (Coreference)} \\
Mult. Entity & \textcolor{darkgray!70}{{\tiny U:} \nlt{Which releases have Motown as their record label?}} \\
& \textcolor{darkgray!70}{{\tiny S:} \nlt{What's Going On, Got to Be There, Can't Slow Down}} \\
& {\tiny U:} \nlt{Which genre do those releases belong to?} \\
\arrayrulecolor{gray!20}\hline
Single Entity (Coreference) & \textcolor{darkgray!70}{{\tiny U:} \nlt{Which narrative location is The Penalty set in?}} \\
& \textcolor{darkgray!70}{{\tiny S:} \nlt{San Francisco}} \\
& {\tiny U:} \nlt{Which color is associated with that film genre?} \\[2ex]
\textbf{Verification (Boolean) (All)  } \\
2 entities, both direct & {\tiny U:} \nlt{Is Zugspitze located in Germany?} \\
\arrayrulecolor{gray!20}\hline
2 entities, one direct and one corefered, object is corefered & \textcolor{darkgray!70}{{\tiny U:} \nlt{Which university was Eden Stiles educated at?}}\\
& \textcolor{darkgray!70}{{\tiny S:} \nlt{University of Michigan}} \\
& \textcolor{darkgray!70}{{\tiny U:} \nlt{And what about C. V. Raman?}} \\
& \textcolor{darkgray!70}{{\tiny S:} \nlt{University of Madras}} \\
& {\tiny U:} \nlt{Was Ravindra Wijegunaratne educated at that university?}\\
\arrayrulecolor{gray!20}\hline
2 entities, one direct and one corefered, subject is corefered & \textcolor{darkgray!70}{{\tiny U:} \nlt{Which German business organization was Gustav Peichl a member of?}} \\
& \textcolor{darkgray!70}{{\tiny S:} \nlt{Academy of Arts, Berlin}} \\
& \textcolor{darkgray!70}{{\tiny U:} \nlt{What was designed by that person?}} \\
& \textcolor{darkgray!70}{{\tiny S:} \nlt{Millennium Tower}} \\
& {\tiny U:} \nlt{Does that tower block belong to Austria?}\\
\arrayrulecolor{gray!20}\hline
3 entities, 2 direct, 2(direct) are query entities, subject is corefered & \textcolor{darkgray!70}{{\tiny U:} \nlt{Which administrative territory was Gary Collier born at?}} \\
& \textcolor{darkgray!70}{{\tiny S:} \nlt{Fort Worth}} \\
& {\tiny U:} \nlt{Is that administrative territory a sister city of Adamsville, New Brunswick and Yuen Long Kau Hui?} \\
\arrayrulecolor{gray!20}\hline
3 entities, all direct, 2 are query entities & {\tiny U:} \nlt{Is Aix-en-Provence partner town of Baton Rouge and Hemmatabad, Alborz?} \\
\arrayrulecolor{gray!20}\hline
one entity, multiple entities (as object) coreferred & \textcolor{darkgray!70}{{\tiny U:} \nlt{Which armed conflicts are Battle of the Argeş or Battle of the Yellow Sea a part of?}} \\
& \textcolor{darkgray!70}{{\tiny S:} \nlt{Romania during World War I, Russo-Japanese War}} \\
& {\tiny U:} \nlt{Did those armed conflicts fight in Rui Natsukawa?} \\[2ex]
\textbf{Quantitative Reasoning (Count) (All) } \\
Incomplete count-based ques & \textcolor{darkgray!70}{{\tiny U:} \nlt{How many people influenced Chris Marker?}} \\
& \textcolor{darkgray!70}{{\tiny S:} \nlt{1}} \\
& \textcolor{darkgray!70}{{\tiny U:} \nlt{And also tell me about Ada Yonath?}} \\
& \textcolor{darkgray!70}{{\tiny S:} \nlt{1}} \\
& {\tiny U:} \nlt{And what about Mikhail Bakunin?}\\
\arrayrulecolor{gray!20}\hline
Count over Atleast/ Atmost/ Approx. the same/Equal|Mult. entity type & {\tiny U:} \nlt{How many cities are the terminus locations of atleast 5 thoroughfares and roads?} \\
\arrayrulecolor{gray!20}\hline
Count over Atleast/ Atmost/ Approx. the same/Equal|Single entity type & {\tiny U:} \nlt{How many musical instruments are played by exactly 2 people?} \\
\arrayrulecolor{gray!20}\hline
Count | Logical operators & {\tiny U:} \nlt{How many bodies of water or watercourses are situated nearby Lübeck?} \\
\arrayrulecolor{gray!20}\hline
Count | Logical operators (Coreference) & \textcolor{darkgray!70}{{\tiny U:} \nlt{Which administrative territory is the native country of Carolina Goic Boroevic?}} \\
& \textcolor{darkgray!70}{{\tiny S:} \nlt{Chile}} \\
& \textcolor{darkgray!70}{{\tiny U:} \nlt{Who is the head of the government of that administrative territory?}} \\
& \textcolor{darkgray!70}{{\tiny S:} \nlt{Michelle Bachelet}} \\
& \textcolor{darkgray!70}{{\tiny U:} \nlt{What is the capital of that administrative territory?}} \\
& \textcolor{darkgray!70}{{\tiny S:} \nlt{Santiago}} \\
& {\tiny U:} \nlt{How many capitals or cities are sister towns of that city?}\\
\arrayrulecolor{gray!20}\hline
Count | Mult. entity type & {\tiny U:} \nlt{How many people starred in Django Kill or Shatterday?} \\
\arrayrulecolor{gray!20}\hline
Count | Single entity type & {\tiny U:} \nlt{How many people starred in Captain America: Civil War?} \\
\arrayrulecolor{gray!20}\hline
Count | Single entity type (Coreference) & \textcolor{darkgray!70}{{\tiny U:} \nlt{Which armed conflict did Lionel of Antwerp, 1st Duke of Clarence take part in?}} \\
& \textcolor{darkgray!70}{{\tiny S:} \nlt{Hundred Years' War}} \\
& {\tiny U:} \nlt{How many people did that armed conflict engage in?} \\[2ex]
\textbf{Comparative Reasoning (Count) (All)} \\
Count over More/Less | Mult. entity type & {\tiny U:} \nlt{How many administrative territories have adopted lesser number of holidays and people as patron saint than Santo Stefano al Mare?} \\
\arrayrulecolor{gray!20}\hline
Count over More/Less | Mult. entity type (Ellipsis) & \textcolor{darkgray!70}{{\tiny U:} \nlt{How many administrative territories have adopted lesser number of holidays and people as patron saint than Santo Stefano al Mare?}} \\
& \textcolor{darkgray!70}{{\tiny S:} \nlt{296}} \\
& {\tiny U:} \nlt{And what about San Donato Milanese?} \\
\arrayrulecolor{gray!20}\hline
Count over More/Less | Mult. entity type (Coreference) & \textcolor{darkgray!70}{{\tiny U:} \nlt{Which administrative territories are Luigi Einaudi the head of state of and have UTC+01:00 as their time zone?}} \\
& \textcolor{darkgray!70}{{\tiny S:} \nlt{Italy}} \\
& {\tiny U:} \nlt{How many administrative territories are the origins of greater number of literary works or releases than that administrative territory?} \\

\arrayrulecolor{gray!20}\hline
Count over More/Less | Single entity type & {\tiny U:} \nlt{How many legislatures represent lesser number of states than East Bengal Legislative Assembly?} \\
\arrayrulecolor{gray!20}\hline
Count over More/Less | Single entity type (Ellipsis) & \textcolor{darkgray!70}{{\tiny U:} \nlt{How many legislatures represent lesser number of states than East Bengal Legislative Assembly?}} \\
& \textcolor{darkgray!70}{{\tiny U:} \nlt{207}} \\
& {\tiny U:} \nlt{And how about Estates of Curaçao?} \\
\arrayrulecolor{gray!20}\hline
Count over More/Less | Single entity type (Coreference) & \textcolor{darkgray!70}{{\tiny U:} \nlt{Which french administrative division was Philippe Esnault born in?}} \\
& \textcolor{darkgray!70}{{\tiny S:}  \nlt{Alençon}} \\
& \textcolor{darkgray!70}{{\tiny U:}  \nlt{Which occupation has that person as his/her 's career?}} \\
& \textcolor{darkgray!70}{{\tiny S:}  \nlt{historian}} \\
& \textcolor{darkgray!70}{{\tiny U:}  \nlt{Which administrative territory is the native country of that person?}} \\
& \textcolor{darkgray!70}{{\tiny S:}  \nlt{France}} \\
& {\tiny U:} \nlt{How many administrative territories inspired less number of fictional locations than that administrative territory?}  \\
\hline
\end{longtable}
\end{center}
}

\twocolumn

\begin{table*}
\begin{tabular}{@{}c@{\hspace{0.2cm}}m{6cm}m{6cm}c@{}}
  & \multicolumn{2}{c}{\textsc{CountLogic}} \\
  \parbox[t]{2mm}{\multirow{4}{*}{\rotatebox[origin=b]{90}{\vspace{9cm}\textsc{{\footnotesize Seen}}\vspace{-1cm}}}} 
  & {\footnotesize Union | Single Relation} & \\
  & \nlutt{Which people are the creators of The Theory of Everything or Ten Minutes to Live?} & 
      \begin{lstlisting}
   SELECT ?x WHERE { 
      {wd:Q15079318 wdt:P162 ?x. ?x wdt:P31 wd:Q502895.} 
      UNION 
      {wd:Q7699260 wdt:P162 ?x. ?x wdt:P31 wd:Q502895.} }
   \end{lstlisting} \\
  & {\footnotesize Count | Single entity type} & \\
  & \nlutt{How many people starred in Captain America: Civil War?} &
   \begin{lstlisting}
   SELECT (COUNT(*) AS ?count) WHERE { 
   wd:Q18407657 wdt:P161 ?x. ?x wdt:P31 wd:Q502895.}
   \end{lstlisting} \\
   \hline
   \parbox[t]{2mm}{\multirow{2}{*}{\rotatebox[origin=b]{90}{\vspace{9cm}\textsc{{\footnotesize UnSeen}}\vspace{-1cm}}}} 
  & {\footnotesize Count | Logical operators} & \\ 
  & \nlutt{How many national association football teams or national sports teams represent Slovenia?} & \begin{lstlisting}
  SELECT (COUNT (DISTINCT ?x) AS ?count) WHERE { 
{?x wdt:P1532 wd:Q215. ?x wdt:P31 wd:Q6979593.} 
UNION 
{?x wdt:P1532 wd:Q215. ?x wdt:P31 wd:Q1194951.} }
  \end{lstlisting} \\
   \hline
   \parbox[t]{2mm}{\multirow{1}{*}{\rotatebox[origin=b]{90}{\textsc{{\footnotesize Pred}}}}} 
   &  
   & \begin{lstlisting}
SELECT (COUNT(DISTINCT ?x) AS ?count) WHERE { 
   {?x wdt:P1532 wd:Q215. ?x wdt:P31 wd: <@\textcolor{red}{Q6979593}@> .} 
   UNION 
   {?x wdt:P1532 wd:Q215. ?x wdt: P31 wd: Q6979593.} }
\end{lstlisting} \\
\hline
\hline
  & \multicolumn{2}{c}{\textsc{UnionMulti}} \\
  \parbox[t]{2mm}{\multirow{4}{*}{\rotatebox[origin=b]{90}{\vspace{9cm}\textsc{{\footnotesize Seen}}\vspace{-1cm}}}} 
  & {\footnotesize Union | Single Relation } & \\
  & \nlutt{Which people are the creators of The Theory of Everything or Ten Minutes to Live?} &
  \begin{lstlisting}
  SELECT ?x WHERE {
     ?x wdt:P915 wd:Q1247373. ?x wdt:P31 wd:Q838948.}
  \end{lstlisting} \\
  & {\footnotesize Count | Mult. entity type} & \\
  & \nlutt{How many people starred in Django Kill or Shatterday?} &
  \begin{lstlisting}
  SELECT (COUNT(DISTINCT ?x) AS ?count) WHERE { 
     {wd:Q1261875 wdt:P161 ?x. ?x wdt:P31 wd:Q502895.} 
     UNION 
     {wd:Q7490688 wdt:P161 ?x. ?x wdt:P31 wd:Q502895.} }
  \end{lstlisting} \\
  \hline
  \parbox[t]{2mm}{\multirow{2}{*}{\rotatebox[origin=b]{90}{\vspace{9cm}\textsc{{\footnotesize UnSeen}}\vspace{-1cm}}}} 
  & {\footnotesize Union | Multiple Relation } & \\
  & \nlutt{Which administrative territories are the origin of Les Chics Types or are the native countries of Robert Kuraś?} &
  \begin{lstlisting}
  SELECT ?x WHERE { 
     {wd:Q3231475 wdt:P495 ?x. ?x wdt:P31 wd:Q15617994.} 
     UNION 
     {wd:Q9310937 wdt:P27 ?x. ?x wdt:P31 wd:Q15617994.} }
  \end{lstlisting} \\
  \hline
  \parbox[t]{2mm}{\multirow{1}{*}{\rotatebox[origin=b]{90}{\vspace{9cm}\textsc{{\footnotesize Pred}}\vspace{6cm}}}} 
  & 
  & \begin{lstlisting}
     SELECT ?x WHERE { 
        {wd:Q3231475 wdt:P495 ?x. ?x wdt:P31 wd:Q15617994.} 
        UNION 
        {wd:Q9310937 wdt:<@\textcolor{red}{P495}@> ?x. ?x wdt:P31 wd:Q15617994.} }
    \end{lstlisting}\\
  \hline
  \hline
  & \multicolumn{2}{c}{\textsc{Verifiy3}} \\
  \parbox[t]{2mm}{\multirow{2}{*}{\rotatebox[origin=b]{90}{\vspace{9cm}\textsc{{\footnotesize Seen}}\vspace{-1cm}}}} 
  & {\footnotesize 2 entities, both direct} & \\
  & \nlutt{Is Zugspitze located in Germany?} &
  \begin{lstlisting}
  ASK {wd:Q3375 wdt:P17 wd:Q183.}
  \end{lstlisting}\\
  \hline
  \parbox[t]{2mm}{\multirow{2}{*}{\rotatebox[origin=b]{90}{\vspace{9cm}\textsc{{\footnotesize UnSeen}}\vspace{-1cm}}}} 
  & {\footnotesize 3 entities, all direct, 2 are query entities} & \\
  & \nlutt{Is Violet Oakley a civilian of United States of America and Scheden?} &  \begin{lstlisting}
  ASK {wd:Q30 wdt:P27 wd:Q1226556. 
       wd:Q557427 wdt:P27 wd:Q1226556.}
  \end{lstlisting}\\
  \hline
  \parbox[t]{2mm}{\multirow{1}{*}{\rotatebox[origin=t]{90}{\textsc{{\footnotesize Pred}}}}} 
  & 
  & \begin{lstlisting}
  ASK {<@\textcolor{red}{wd:Q1226556 wdt:P27 wd:Q30}@>. 
       wd:Q557427 wdt:P27 wd:<@\textcolor{red}{Q557427}@>.}
 \end{lstlisting}
 \end{tabular} 
\caption{Generalisation splits, unseen question sub-types, support
question sub-types seen during training, and example common errors on unseen
predictions.}\label{tab:generalisation:errors}
\end{table*}

\end{document}